%% file: dpbp-arxiv-2015.tex
\documentclass[11pt]{article}
\pdfoutput=1  %

\usepackage{acl2012}
\usepackage{times}
\usepackage{latexsym}
\usepackage{amsmath}
\usepackage{multirow}
\usepackage{url}

\usepackage{amssymb}
\usepackage{amsfonts}
\usepackage{graphicx}
\usepackage{booktabs} %
\usepackage{verbatim} %
\usepackage{tabu}
\usepackage{rotating} %
\usepackage{makecell}
\usepackage{pict2e} %
\usepackage{paralist}
\usepackage[percent]{overpic}

\usepackage{soul} %
\usepackage[usenames,dvipsnames,svgnames,table]{xcolor} %

\usepackage{arydshln} %

\usepackage{tikz}
\usetikzlibrary{shapes,arrows}
\usetikzlibrary{automata}
\usetikzlibrary{positioning}
\usetikzlibrary{arrows,shapes,automata,backgrounds,petri,positioning}
\usetikzlibrary{decorations.pathmorphing}
\usetikzlibrary{decorations.shapes}
\usetikzlibrary{decorations.text}
\usetikzlibrary{decorations.fractals}
\usetikzlibrary{decorations.footprints}
\usetikzlibrary{shadows}
\usetikzlibrary{calc}
\usetikzlibrary{spy}

\makeatletter
\newcommand{\@BIBLABEL}{\@emptybiblabel}
\newcommand{\@emptybiblabel}[1]{}
\makeatother
\usepackage[hidelinks]{hyperref}

\usepackage[disable]{todonotes}   %

\newcommand{\vc}[1]{\boldsymbol{#1}}

\DeclareMathOperator*{\argmax}{argmax}
\DeclareMathOperator*{\argmin}{argmin}

\newcommand{\tbhead}[1]{\textsc{#1}}
\newcommand{\CommentForSpace}[1]{}

\newcommand{\ON}{\textsc{on}}
\newcommand{\OFF}{\textsc{off}}

\newcommand{\adj}[1]{\eth #1}
\newcommand{\deriv}[2]{\tfrac{\partial #1}{\partial #2}}

\renewcommand{\adj}[1]{\eth #1}

\newcommand{\yvar}[1]{y_{#1}}
\newcommand{\yalphavar}[1]{\vc{y}_{\alpha} \sim \yvar{#1}}
\newcommand{\msg}{m}
\newcommand{\domain}[1]{}
\newcommand{\Nalpha}{\mathcal{N}(\alpha)}
\newcommand{\Ni}{\mathcal{N}(i)}
\newcommand{\PTree}{\textsc{PTree}}
\newcommand{\Tree}{\textsc{Tree}}
\newcommand{\vtheta}{\vc{\theta}}

\title{Approximation-Aware Dependency Parsing by Belief Propagation}

\author{Matthew R. Gormley \qquad  Mark Dredze \qquad Jason Eisner \\
Department of Computer Science\\ 
Center for Language and Speech Processing\\
Human Language Technology Center of Excellence\\
Johns Hopkins University, Baltimore, MD\\
\tt{\{mrg,mdredze,jason\}@cs.jhu.edu}}

\date{}

\begin{document}
\maketitle
\begin{abstract}
  We show how to train the fast dependency parser of
  \newcite{smith_eisner_2008_bp} for improved accuracy.  This parser
  can consider higher-order interactions among edges while retaining
  $O(n^3)$ runtime.  It outputs the parse with maximum expected
  recall---but for speed, this expectation is taken under a posterior
  distribution that is constructed only approximately, using loopy
  belief propagation through structured factors.  We show how to
  adjust the model parameters to compensate for the errors introduced
  by this approximation, by following the gradient of the actual loss
  on training data.  We find this gradient by back-propagation.  That
  is, we treat the entire parser (approximations and all) as a
  differentiable circuit, as \newcite{stoyanov_empirical_2011}
  {} and \newcite{domke-2010}
  did for loopy CRFs.
  The resulting trained parser obtains higher accuracy with fewer iterations
  of belief propagation than one trained by conditional log-likelihood.
  {}
\end{abstract}

\section{Introduction}
\label{sec:intro}

Recent improvements to dependency parsing accuracy have been driven by
higher-order features.  Such a feature can look beyond just the parent and
child words connected by a single edge to also consider siblings, grand-parents, etc.
By including increasingly global information, these features
provide more information for the parser---but they also complicate inference.
The resulting higher-order parsers depend on
{\em approximate} inference and decoding procedures,
which may prevent them from predicting the best
parse.

For example, consider the dependency parser we will train in this paper,
which is based on the work of \newcite{smith_eisner_2008_bp}.
Ostensibly, this parser finds the minimum Bayes risk (MBR) parse under a
probability distribution defined by a higher-order dependency parsing
model.  In reality, however, it achieves $O(n^3T)$ runtime by relying
on three \emph{approximations during inference}: (1) variational
inference by loopy belief propagation (BP) on a factor graph, (2)
early stopping of inference after $t_{\textrm{max}}$ iterations prior to convergence, and (3) a
first-order pruning model to limit the number of edges considered in
the higher-order model.  Such parsers are traditionally trained {\em
  as if the inference had been exact} \cite{smith_eisner_2008_bp}.\footnote{For perceptron
  training, utilizing inexact inference as a drop-in replacement
  for exact inference can badly mislead the learner \cite{kulesza_structured_2008}.}
{}  

In contrast, we train the parser such that the \emph{approximate}
system performs well on the final evaluation function.
\newcite{stoyanov_minimum-risk_2012} call this approach ERMA, for ``empirical
  risk minimization under approximations.''
We treat the entire parsing computation as a differentiable circuit, and backpropagate the evaluation function 
{}
through our approximate inference and decoding methods to
improve its parameters by gradient descent. {}

Our primary contribution is
the application of Stoyanov and Eisner's learning method in the
parsing setting, for which the graphical model involves a \emph{global constraint}.
\newcite{smith_eisner_2008_bp} previously showed how to 
run BP in this setting (by calling the inside-outside
algorithm as a subroutine).  We must backpropagate the downstream objective function 
through their algorithm so that we can follow its gradient.
{}
We carefully define our objective function to be smooth and
differentiable, yet equivalent to accuracy of the minimum Bayes risk
(MBR) parse in the limit. Further we introduce a new simpler objective
function based on the L$_2$ distance between the approximate marginals
and the ``true'' marginals from the gold data.

The goal of this work is to account for the approximations made by a
system rooted in structured belief propagation. Taking such
approximations into account during training enables us to improve the
speed and accuracy of inference at test time. To this end, we compare
our training method with the standard approach of conditional
log-likelihood. We evaluate our parser
on 19 languages from the CoNLL-2006 \cite{buchholz_conll-x_2006} and
CoNLL-2007 \cite{nivre_conll_2007} Shared Tasks as well as the English
Penn Treebank \cite{marcus_building_1993}. On English, the resulting
parser obtains higher accuracy with fewer iterations of BP than
standard conditional log-likelihood (CLL) training. On the CoNLL
languages, we find that on average it yields higher accuracy parsers
than CLL training, particularly when limited to few BP iterations.

\section{Dependency Parsing by Belief Propagation}
\label{sec:dpbybp}

This section describes the parser that we will train.

\paragraph{Model} 

\begin{figure}[tb]
\centering
\includegraphics[width=\linewidth]{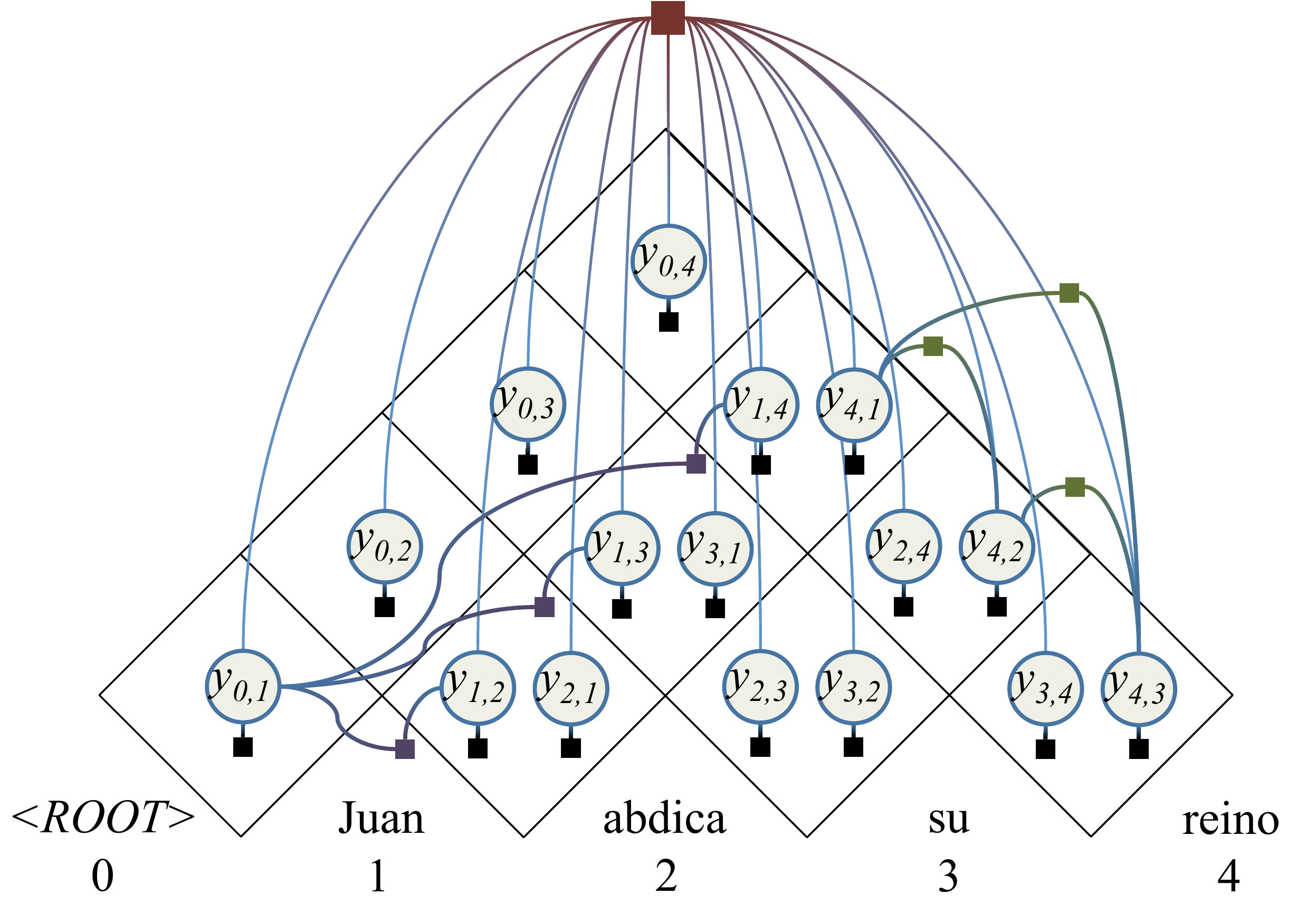}
\caption{Factor graph for dependency parsing of a 4-word sentence;
  the special node {\em $<$ROOT$>$} is the root of the dependency graph.
  In this figure, the boolean variable $y_{h,m}$ encodes whether 
  the edge from parent $h$ to child $m$ is present.  The unary
  factor (black) connected to this variable scores the edge in
  isolation (given the sentence).  The \PTree{} factor (red)
  coordinates all variables to ensure that the edges form a tree.
  The drawing shows a few of the higher-order factors (purple factors
  for grandparents, green factors for arbitrary siblings); these
  are responsible for the graph being cyclic (``loopy'').}
\label{fig:dpFactorGraph}
\end{figure}

{}

{}

A factor graph \cite{frey_factor_1997,kschischang_factor_2001} is a
bipartite graph between factors $\alpha$ and variables $y_i$, and defines
the factorization of a probability distribution over a set of
variables $\{y_1, y_2, \ldots\}$.  The factor graph contains edges
between each factor $\alpha$ and a subset of variables
$\vc{y}_{\alpha}$. Each factor has a local opinion about the possible
assignments to its neighboring variables. Such opinions are given by
the factor's potential function $\psi_{\alpha}$, which assigns a
nonnegative score to each configuration of a subset of variables
$\vc{y}_{\alpha}$.
We define the probability of a given assignment $\vc{y}$ to be
proportional to a product of potential functions: $p(\vc{y}) =
\frac{1}{Z} \prod_{\alpha} \psi_{\alpha}(\vc{y}_{\alpha})$.  

\newcite{smith_eisner_2008_bp} define a factor graph for dependency
parsing of a given $n$-word sentence: $n^2$ binary variables
$\{y_1, y_2, \ldots\}$ {} indicate which of
the directed arcs are included ($y_i = \ON$) or excluded ($y_i = \OFF$) in the dependency parse. 
One of the factors plays the role of a hard global constraint:
$\psi_{\PTree}(\vc{y})$ is 1 or 0 according to whether the assignment
encodes a projective dependency tree.  Another $O(n^2)$ factors (one
per variable) evaluate the individual arcs given the sentence, so that
$p(\vc{y})$ describes a first-order dependency parser.  A higher-order
parsing model is achieved by including higher-order factors, each scoring
configurations of two or more arcs, such as grandparent and sibling
configurations. Higher-order factors add cycles to the
factor graph. 
See Figure
\ref{fig:dpFactorGraph} for an example factor graph. 

We define each
potential function to have a log-linear form:
$\psi_{\alpha}(\vc{y}_{\alpha}) = \exp(\vtheta \cdot
\vc{f}_{\alpha}(\vc{y}_{\alpha},\vc{x}))$.  Here $\vc{x}$ is the
vector of observed variables such as the sentence and its POS tags; $\vc{f}_{\alpha}$ extracts a vector of
features; and $\vtheta$ is our vector of model parameters.  We write the
resulting probability distribution over parses as $p_{\vtheta}(\vc{y})$,
to indicate that it depends on $\vtheta$.

\paragraph{Loss}
For dependency parsing, our loss function is the number of missing
edges in the predicted parse $\hat{\vc{y}}$, relative to the reference (or ``gold'') parse $\vc{y}^*$:
\begin{align} 
  \ell(\hat{\vc{y}}, \vc{y}^*) =  \sum_{i:\, y_i^* = \ON} \delta(\hat{y}_i = \OFF) \label{eq:directedDependencyError}
\end{align}
Because $\hat{\vc{y}}$ and $\vc{y}^*$ each specify exactly one parent for each
word token, $\ell(\hat{\vc{y}}, \vc{y}^*)$ equals the number of word tokens
whose parent is predicted incorrectly---that is, directed dependency error.

\paragraph{Decoder}

To obtain a single parse as output, we use a minimum Bayes risk (MBR)
decoder, which attempts to find the tree with minimum expected loss
under the model's distribution
\cite{bickel_mathematical_1977}. For our directed dependency error
loss function, we obtain the following decision rule:
\begin{align}
  h_{\vtheta}(\vc{x}) &= \argmin_{\hat{\vc{y}}} \;\mathbb{E}_{p_{\vtheta}(\vc{y} | \vc{x})}[\ell(\hat{\vc{y}}, \vc{y})] \label{eq:mbrGeneral}\\
                           &= \argmax_{\hat{\vc{y}}} \sum_{i:\, \hat{y}_i = \ON} p_{\vtheta}(y_i = \ON | \vc{x}) \label{eq:mbrDecoding} 
\end{align}
Here $\hat{\vc{y}}$ ranges over well-formed parses.  Thus, our parser
seeks a well-formed parse $h_{\vtheta}(\vc{x})$ whose individual edges
have a high probability of being correct according to $p_{\vtheta}$.  MBR
is the principled way to take a loss function into account under a
probabilistic model.  By contrast, maximum \emph{a posteriori} (MAP)
decoding does not consider the loss function.  It would return the
single highest-probability parse even if that parse, and its
individual edges, were unlikely to be correct.\footnote{If we used a
  simple 0-1 loss function within \eqref{eq:mbrGeneral}, then MBR
  decoding would reduce to MAP decoding.}  

All systems in this paper use MBR decoding to consider the loss
function at {\em test} time.  This implies that the ideal training procedure
would be to find the {\em true} $p_{\vtheta}$ so that its marginals can
be used in \eqref{eq:mbrDecoding}.  Our baseline system attempts this.
In practice, however, we will not
be able to find the true $p_{\vtheta}$ (model misspecification) nor
exactly compute the marginals of $p_{\vtheta}$ (computational intractability).  Thus,
this paper proposes a training procedure that compensates for the system's
approximations, adjusting $\vtheta$  to reduce the
actual loss of $h_{\vtheta}(\vc{x})$ as measured at {\em training} time.

{}
{}
{}

To find the MBR parse, we first run inference to compute the
marginal probability $p_{\vtheta}(y_i=\ON)$ for each edge.
Then we maximize \eqref{eq:mbrDecoding} by running a first-order dependency parser with edge scores
equal to those probabilities.\footnote{Prior work
  \cite{smith_eisner_2008_bp,bansal_structured_2014} 
  used the log-odds ratio $\log \frac{p_{\vtheta}(y_i=\ON)}{p_{\vtheta}(y_i=\OFF)}$
  as the edge scores for decoding, but this yields a parse different from the
  MBR parse.}
When our inference algorithm is approximate, 
we replace the exact marginal with
its approximation---the normalized belief from BP, given by
$b_i(\ON)$ in \eqref{eq:beliefVar} below.

\paragraph{Inference}

Loopy belief propagation (BP) \cite{murphy_loopy_1999} computes
approximations to the variable marginals $p_{\vtheta}(y_i)$ and
the factor marginals $p_{\vtheta}(\vc{y}_{\alpha})$.
{} 
{} 
The algorithm proceeds by iteratively sending
messages from variables, $y_i$, to factors, $\psi_{\alpha}$:
\begin{align}
\msg^{(t)}_{i \rightarrow \alpha}(y_i) = \prod_{\beta \in \Ni \backslash \alpha} \msg^{(t-1)}_{\beta \rightarrow i}(y_i) 
\label{eq:msgVarToFac}
\end{align}
\noindent and from factors to variables:
\begin{align}
\msg^{(t)}_{\alpha \rightarrow i}(y_i) = \sum_{\yalphavar{i}}  \psi_{\alpha}(\vc{y}_{\alpha}) \prod_{j \in \Nalpha \backslash i} \msg^{(t-1)}_{j \rightarrow \alpha}(\yvar{i}) 
\raisetag{12pt}
\label{eq:msgFacToVar}
\end{align}
\noindent where $\Ni$ and $\Nalpha$ denote the neighbors of $y_i$ and $\psi_{\alpha}$
respectively, and where $\yalphavar{i}$ is standard notation
to indicate that $\vc{y}_{\alpha}$ ranges over all assignments to the
variables participating in the factor $\alpha$ provided that the
$i$th variable has value $y_i$. Note that the messages at
time $t$ are computed from those at time $(t-1)$.
Messages at the final time $t_{\textrm{max}}$ are used to compute the {\em
  beliefs} at each factor and variable:
\begin{align}
&b_i(y_i) = \prod_{\alpha \in \Ni} \msg^{(t_{\textrm{max}})}_{\alpha \rightarrow i}(y_i) \label{eq:beliefVar}\\
&b_{\alpha}(\vc{y}_{\alpha}) = \psi_{\alpha}(\vc{y}_{\alpha}) \prod_{i \in \Nalpha} \msg^{(t_{\textrm{max}})}_{i \rightarrow \alpha}(\yvar{i}) \label{eq:beliefFac}
\end{align}
{} 
Each of the functions defined by equations
\eqref{eq:msgVarToFac}--\eqref{eq:beliefFac} can be optionally
rescaled by a constant at any time, e.g., to prevent
overflow/underflow.  Below, we specifically assume that each function
$b_i$ has been rescaled such that $\sum_{y_i} b_i(y_i)=1$.  This $b_i$
approximates the marginal distribution over $y_i$ values.
{}

Messages continue to change indefinitely if the factor graph is
cyclic, but in the limit, the rescaled messages may converge.
Although the equations above update all messages in parallel,
convergence is much faster if only one message is updated per
timestep, in some well-chosen {\em serial} order.\footnote{Following
  \newcite[footnote 22]{dreyer_graphical_2009}, we choose an arbitrary
  directed spanning tree rooted at the \PTree{} factor.  We visit the
  nodes in topologically sorted order (starting at the leaves) and
  update any message from the node being visited to a node that is
  {\em later} in the order (e.g., closer to the root).  We then
  reverse this order and repeat, so that every message has been passed
  once.  This constitutes one {\bf iteration} of BP.}

For the \PTree{} factor, the summation over variable
assignments required for $\msg^{(t)}_{\alpha \rightarrow i}(y_i)$ in
Eq. \eqref{eq:msgFacToVar} equates to a
summation over exponentially many projective parse trees. However, we can use
an inside-outside variant of the algorithm of 
\newcite{eisner_three_1996} to compute this in polynomial time (we describe
this as hypergraph parsing in \S~\ref{sec:learning}).  
The resulting ``structured BP'' inference
procedure is exact for first-order dependency parsing, and approximate 
when high-order factors are incorporated. The advantage of BP is that 
it enables fast approximate inference when exact inference is too
slow. See \newcite{smith_eisner_2008_bp} for details.\footnote{How
  slow is exact inference for dependency parsing?  For certain choices
of higher-order factors, polynomial time is possible via dynamic programming
\cite{mcdonald_online_2005,carreras_experiments_2007,koo_efficient_2010}.  However, 
BP will typically be asymptotically faster (for a fixed number of iterations) and faster in practice.
In some other settings, exact inference is NP-hard.  In particular,
non-projective parsing becomes NP-hard with even second-order factors
\cite{mcdonald_online_2006}.  BP can handle this case in polynomial time 
by replacing the \PTree{} factor with a \Tree{} factor that allows edges to cross.}

\section{Approximation-aware Learning}
\label{sec:learning}

We aim to find the parameters $\vtheta^*$ that minimize a regularized
objective function over the training sample of sentence/parse pairs
$\{(\vc{x}^{(d)}, \vc{y}^{(d)})\}_{d=1}^D$. 
\begin{align}\label{eq:regemprisk}
\vtheta^* = \argmin_{\vtheta} \frac{\lambda}{2} ||\vtheta||_2^2 + \frac{1}{D} \sum_{d=1}^D J(\vtheta; \vc{x}^{(d)}, \vc{y}^{(d)}) 
\raisetag{12pt}
\end{align}
where $\lambda > 0$ is the regularization coefficient and $J(\vtheta; \vc{x},
\vc{y})$ is a given differentiable function, possibly nonconvex.
We locally minimize this objective using $\ell{}_2$-regularized
AdaGrad with Composite Mirror Descent 
\cite{duchi_adaptive_2011}---a variant of stochastic gradient descent
that uses mini-batches, an adaptive learning rate per dimension, and sparse lazy
updates from the regularizer.\footnote{$\vtheta$ is
  initialized to $\vc{0}$ when not otherwise specified.}

\paragraph{Objective Functions}
As in \newcite{stoyanov_empirical_2011}, our aim is to minimize
expected loss on the true data distribution over sentence/parse pairs
$(X,Y)$:
\begin{align} \textstyle
  \vtheta^* = \argmin_{\vtheta}\,\mathbb{E}[\ell(h_{\vtheta}(X), Y)] \label{eq:expectedLoss}
\end{align}
Since the true data distribution is unknown, we substitute
the expected loss over the training sample, and regularize 
our objective to reduce sampling variance.  Specifically, we aim to minimize the \textbf{regularized empirical
  risk}, given by \eqref{eq:regemprisk} with $J(\vtheta; \vc{x}^{(d)}, \vc{y}^{(d)})$ set to $\ell(h_{\vtheta}(\vc{x}^{(d)}),\vc{y}^{(d)})$.  
Using our MBR decoder $h_{\vtheta}$ in \eqref{eq:mbrDecoding},
this loss function would not be differentiable because of
the $\argmax$ in the definition of $h_{\vtheta}$
\eqref{eq:mbrDecoding}. We will address this below by substituting a
differentiable softmax.  This is the ``ERMA'' method of \newcite{stoyanov_minimum-risk_2012}.
We will also consider simpler choices of $J(\vtheta;
\vc{x}^{(d)}, \vc{y}^{(d)})$ that are commonly used in training neural networks.
Finally, the standard convex objective is conditional log-likelihood (\S~\ref{sec:cll}).

\paragraph{Gradient Computation}
To compute the gradient $\nabla_{\vtheta} J(\vtheta; \vc{x},
\vc{y}^*)$ of the loss on a single sentence
$(\vc{x}, \vc{y}^*) = (\vc{x}^{(d)}, \vc{y}^{(d)})$, we
apply automatic differentiation (AD) in the reverse mode
\cite{griewank_automatic_1991}.  This yields the same type of
``back-propagation'' algorithm that has long been used for training
neural networks \cite{rumelhart-hinton-williams-1986}. 
In effect, we are
regarding (say) 5 iterations of the BP algorithm on sentence $\vc{x}$,
followed by (softened) MBR decoding and comparison to the target output $\vc{y}^*$, as
a kind of neural network that computes $\ell(h_{\vtheta}(\vc{x}),\vc{y}^*)$.  It is important to note that the resulting
gradient computation algorithm is exact up to floating-point error, and has the same asymptotic
complexity as the original decoding algorithm, requiring only about
twice the computation.  The AD method applies provided that the original
function is indeed differentiable with respect to $\vtheta$, an issue
that we take up below.

In principle, it is possible to compute the gradient with minimal
additional coding.  There exists AD software (some listed at
\url{autodiff.org}) that could be used to derive the
necessary code automatically. Another option would be to use the
perturbation method of \newcite{domke-2010}.  However, we implemented
the gradient computation directly, and we describe it here.

\subsection{Inference, Decoding, and Loss as a Feedfoward Circuit}
\label{sec:feedforward}

The backpropagation algorithm is often applied to neural networks,
where the topology of a feedforward circuit is statically specified
and can be applied to any input. Our BP algorithm, decoder, and loss
function similarly define a feedfoward circuit that computes our
function $J$.  However, the circuit's topology is defined
\emph{dynamically} (per sentence $\vc{x}^{(d)}$) by ``unrolling'' the computation
into a graph.

\begin{figure}[tb]
  \resizebox{\textwidth/2}{!}{
    \input{feedforward.tex}
  }%
  \caption{Feed-forward topology of inference, decoding, and loss. (E)
  shows the \emph{annealed risk}, one of the objective
  functions we consider.}
  \label{fig:feedforward}
\end{figure}
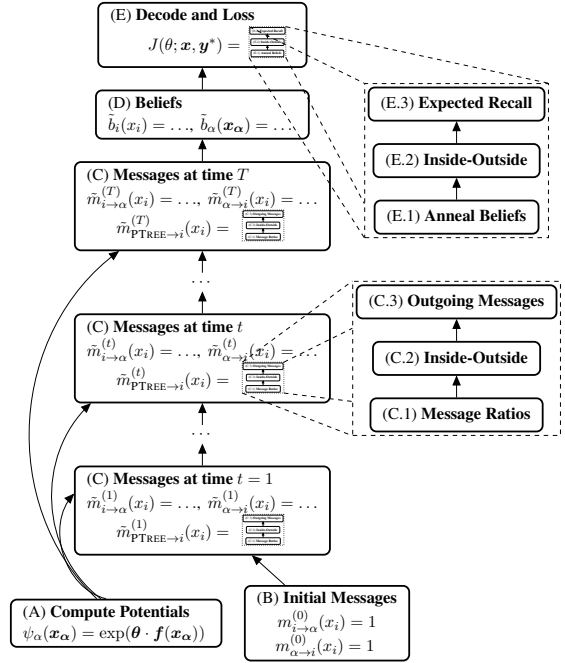

Figure \ref{fig:feedforward} shows this topology for one choice of
objective function. The high level modules consist of (A) computing potential functions, (B) initializing
messages, (C) sending messages, (D) computing beliefs, and (E) decoding
and computing the loss. 
We \emph{zoom in} on two submodules: the
first computes messages from the \PTree{} factor efficiently (C.1--C.3); the
second computes a softened version of our loss function (E.1--E.3). Both of these
submodules are made efficient by the inside-outside
algorithm. 

The remainder of this section describes additional details of how we
define the function $J$ (the forward pass) and how we compute its
gradient (the backward pass).
Backpropagation computes the derivative of any given function
specified by an arbitrary circuit
consisting of elementary differentiable operations (e.g. $+, -,
\times, \div, \log, \exp$). This is accomplished by repeated
application of the chain rule.
\CommentForSpace{$\deriv{y}{x} =
\deriv{y}{u}\deriv{u}{x}$. {}
If one were to compute
$\deriv{y}{x}$ for the equation $y = ax^2 + b(\exp(x)) /
cx$, the first step would be to determine the intermediate quantities
$u$ for application of the chain rule.}

Backpropagating through an \emph{algorithm} proceeds by similar application
of the chain rule, where the intermediate quantities are determined by
the topology of the circuit. Doing so with the circuit from Figure
\ref{fig:feedforward} poses several
challenges. \newcite{eaton_choosing_2009} and
\newcite{stoyanov_empirical_2011} showed how to
backpropagate through the basic BP algorithm, and we reiterate the key details
below (\S~\ref{sec:backbp}).  The remaining challenges form the primary
technical contribution of this paper:
\begin{compactenum}
\item Our true loss function $\ell(h_{\vtheta}(\vc{x}),\vc{y}^*)$ by
  way of the decoder \eqref{eq:mbrDecoding} contains an $\argmax$ over
  trees and is therefore not differentiable. We show how to soften
  this decoder, making it differentiable (\S~\ref{sec:backloss}).
\item Empirically, we find the above objective difficult to
  optimize. To address this, we substitute a simpler L$_2$ loss function
  (commonly used in neural networks).  
  This is easier to optimize and yields
  our best parsers in practice (\S~\ref{sec:backloss}).
\item We show how to run backprop through the inside-outside algorithm
  on a hypergraph (\S~\ref{sec:backinsideoutside}), and thereby on the
  softened decoder and computation of messages from the \PTree{}
  factor. This allows us to go beyond
  \newcite{stoyanov_empirical_2011} and train \emph{structured} BP in
  an approximation-aware and loss-aware fashion.
\end{compactenum}

\vspace{1em} 

\subsection{Differentiable Objective Functions}
\label{sec:backloss}

\newcommand*\rfrac[2]{#1/#2} 

\paragraph{Annealed Risk}

{}

Directed dependency error, $\ell(h_{\vtheta}(\vc{x}), \vc{y}^*)$, is
not differentiable due to the $\argmax$ in the decoder
$h_{\vtheta}$. We therefore redefine $J(\vtheta; \vc{x},
\vc{y}^*)$ to be a new \emph{differentiable} loss
function, the \textbf{annealed risk}
$R_{\vtheta}^{\rfrac{1}{T}}(\vc{x}, \vc{y}^*)$, which approaches the
loss $\ell(h_{\vtheta}(\vc{x}), \vc{y}^*)$ as the \textbf{temperature}
$T\rightarrow 0$.

This is done by replacing our non-differentiable decoder
$h_{\vtheta}$ with a differentiable one (at training time).  As input, it still takes the 
marginals $p_{\vtheta}(y_i=\ON\mid\vc{x})$, or in practice, their BP approximations $b_i(\ON)$.  We
define a distribution over parse trees:
\begin{align} \label{eq:stochDecoder}
\!q_{\vtheta}^{\rfrac{1}{T}}(\hat{\vc{y}}) \propto \exp \left( \sum_{i : \hat{y}_i \text{=} \ON} p_{\vtheta}(y_i=\ON|\vc{x}) / T \right)
\end{align}
Imagine that at training time, our decoder stochastically returns a
parse $\hat{\vc{y}}$ {\em sampled} from this distribution.
Our risk is the expected loss of that decoder:
\begin{align}
R_{\vtheta}^{\rfrac{1}{T}}(\vc{x}, \vc{y}^*) &=
\mathbb{E}_{\hat{\vc{y}} \sim q_{\vtheta}^{\rfrac{1}{T}}}[\ell(\hat{\vc{y}}, \vc{y}^*)] 
\end{align}
As $T\rightarrow 0$ (``annealing''), the decoder almost always chooses the MBR
parse,\footnote{Recall from
\eqref{eq:mbrDecoding} that the MBR parse is
the tree $\hat{\vc{y}}$ that maximizes the sum $\sum_{i : \hat{y}_i = \ON}
p_{\vtheta}(y_i=\ON|\vc{x})$. As $T \rightarrow 0$, the right-hand
side of \eqref{eq:stochDecoder} grows fastest for this
$\hat{\vc{y}}$, so its probability under $q_{\vtheta}^{\rfrac{1}{T}}$
approaches 1 (or $1/k$ if there is a $k$-way tie for MBR parse).}
so our risk approaches the loss of the actual MBR
decoder that will be used at test time.  However, as a function of
$\vc{\theta}$, it remains differentiable (though not convex) for any $T>0$.

To compute the annealed risk, observe that it simplifies to
$R_{\vtheta}^{\rfrac{1}{T}}(\vc{x}, \vc{y}^*) = -\sum_{i : y_i^* = \ON}
q_{\vtheta}^{\rfrac{1}{T}}(\hat{y}_i=\ON)$. This is the negated \textbf{expected recall}
of a parse $\hat{\vc{y}} \sim q_{\vtheta}^{\rfrac{1}{T}}$. We obtain
the required marginals 
$q_{\vtheta}^{\rfrac{1}{T}}(\hat{y}_i=\ON)$ from
\eqref{eq:stochDecoder} by running
inside-outside where the edge weight for edge $i$ is given by
$\exp(p_{\vtheta}(y_i=\ON|\vc{x}) / T)$. {} 

With the annealed risk as our $J$ function, we can compute
$\nabla_{\vc{\theta}} J$ by backpropagating through the computation in
the previous paragraph.  The computations of the edge weights and the
expected recall are trivially differentiable.  The only challenge is
computing the partials of the marginals 
differentiating the function computed by this call to the inside-outside
algorithm; we address this in Section
\ref{sec:backinsideoutside}. Figure \ref{fig:feedforward} (E.1--E.3)
shows where these computations lie within the circuit.

Whether our test-time system computes the marginals of $p_{\vtheta}$ 
exactly or does so approximately via BP, our new training objective approaches
(as $T\rightarrow 0$) the
true empirical risk of the test-time parser that performs MBR decoding from
the computed marginals.
Empirically, however, we will find that it is not the most effective training
objective (\S~\ref{sec:results}). \newcite{stoyanov_empirical_2011} postulate that the
nonconvexity of empirical risk may make it a
difficult function to optimize (even with annealing). Our next two objectives provide
alternatives.

\paragraph{L$_2$ Distance}
We can view our inference, decoder, and loss as defining a form of
deep neural network, whose topology is inspired by our linguistic
knowledge of the problem (e.g., the edge variables should define a tree). This
connection to deep learning allows us to consider training methods
akin to supervised layer-wise training.
We temporarily remove the top layers of our network (i.e. the decoder
and loss module, Fig.~\ref{fig:feedforward} (E)) so that the output
layer of our ``deep network'' consists of the normalized variable beliefs
$b_i(y_i)$ from BP. We can then define a supervised loss function directly on
these beliefs. 

We don't have supervised data for this layer of beliefs, but we can
create it artificially. Use the supervised parse $\vc{y}^*$ to define
``target beliefs'' by $b^*_i(y_i) = \mathbb{I} 
(y_i=y^*_i) \in \{0,1\}$.  
To find parameters $\vc{\theta}$ that make BP's beliefs close to 
these targets, we can minimize an \textbf{L$_2$
  distance} loss function:
\begin{align}
 J(\vtheta; \vc{x}, \vc{y}^*) = \sum_i \sum_{y_i} (b_i(y_i) - b_i^*(y_i^*))^2 \label{eq:l2loss}
\end{align}
We can use this L$_2$ distance objective function for training, adding
the MBR decoder and loss evaluation back in only at test time.

\paragraph{Layer-wise Training}
Just as in layer-wise training of neural networks, we can take a two-stage approach to
training.  First, we train to minimize the L$_2$ distance. Then, we use
the resulting $\vc{\theta}$ as initialization to optimize the annealed
risk, which does consider the decoder and loss function (i.e. the top
layers of Fig.~\ref{fig:feedforward}).
\newcite{stoyanov_empirical_2011} found mean squared error (MSE) to 
give a smoother training objective, though still
non-convex, and similarly used it to find an initializer for empirical risk. Though
their variant of the L$_2$ objective did not completely dispense with
the decoder as ours does, it is a similar approach to our proposed layer-wise training.

\subsection{Backpropagation through BP}
\label{sec:backbp}

Belief propagation proceeds iteratively by sending messages. We can
label each message with a timestamp $t$ (e.g. $\msg_{i \rightarrow
  \alpha}^{(t)}$) indicating the time step at which it was
computed. Figure \ref{fig:feedforward} (B) shows the messages at time
$t=0$, denoted $\msg_{i \rightarrow \alpha}^{(0)}$, which are
initialized to the uniform distribution.  Figure \ref{fig:feedforward}
(C) depicts the computation of all subsequent messages via Eqs.
\eqref{eq:msgVarToFac} and \eqref{eq:msgFacToVar}. Messages at time
$t$ are computed from messages at time $t-1$ or before and the
potential functions $\psi_{\alpha}$.  After the final iteration $T$,
the beliefs $b_i(y_i), b_{\alpha}(\vc{y}_{\alpha})$ are computed from
the final messages $\msg_{i \rightarrow \alpha}^{(T)}$ using Eqs.
\eqref{eq:beliefVar} and \eqref{eq:beliefFac}---this is shown in
Figure \ref{fig:feedforward} (D). Optionally, we can normalize the
messages after each step to avoid overflow (not shown in the figure)
as well as the beliefs.

Except for the messages sent from the \PTree{} factor, each step of BP
computes some value from earlier values using a {\em simple} formula.
Back-propagation differentiates these simple formulas.  This lets it
compute $J$'s partial derivatives with respect to the earlier values,
once its partial derivatives have been computed with respect to later
values.  Explicit formulas can be found in the appendix of
\newcite{stoyanov_empirical_2011}.
{}

\subsection{BP and backpropagation with \PTree{}}
\label{sec:backptree}

The \PTree{} factor has a special structure that we exploit for
efficiency during BP.
\newcite{stoyanov_empirical_2011} assume that BP takes an {\em
  explicit} sum in \eqref{eq:msgFacToVar}.  For the \PTree{} factor,
this equates to a sum over all projective dependency trees (since
$\psi_{\PTree{}}(\vc{y})=0$ for any assignment $\vc{y}$ which is not a
tree).  There are exponentially many such trees.  However,
\newcite{smith_eisner_2008_bp} point out that for $\alpha=\PTree$, the
summation has a special structure that can be exploited by dynamic
programming.

To compute the factor-to-variable messages from $\alpha=\PTree$, they first
run the inside-outside algorithm where the edge weights are given by
the ratios of the messages to \PTree{}:
 $ \frac{\msg_{i \rightarrow \alpha}^{(t)}(\ON)}{\msg_{i \rightarrow \alpha}^{(t)}(\OFF)}$.
Then they multiply each resulting edge marginal given by inside-outside
by the product of all the $\OFF$ messages $\prod_i \msg_{i \rightarrow
  \alpha}^{(t)}(\OFF)$ to get the marginal factor belief $b_{\alpha}(y_i)$. Finally
they divide the belief by the incoming message $\msg_{i \rightarrow
  \alpha}^{(t)}(\ON)$ to get the corresponding outgoing message
$\msg_{\alpha \rightarrow i}^{(t+1)}(\ON)$. 

These steps are shown in Figure \ref{fig:feedforward} (C.1--C.3), and
are repeated each time we send a message from the PTree{} factor. The
derivatives of the message ratios and products mentioned here are
trivial.  Though we focus here on projective dependency parsing, our
techniques are also applicable to non-projective parsing and the
\Tree{} factor; we leave this to future work. In the next subsection,
we explain how to backpropagate through the inside-outside algorithm.

\subsection{Backpropagation through Inside-Outside on a
  Hypergraph}
\label{sec:backinsideoutside}

Both the annealed risk loss function (\S~\ref{sec:backloss}) and the
computation of messages from the \PTree{} factor use the
inside-outside algorithm for dependency parsing. Here we describe
inside-outside and the accompanying backpropagation algorithm over a
hypergraph. This more general treatment shows the applicability of our
method to other structured factors such as for CNF parsing, HMM
forward-backward, etc. In the case of dependency parsing, the
structure of the hypergraph is given by the dynamic programming
algorithm of \newcite{eisner_three_1996}. {}
{}

For the {\bf forward pass} of the
inside-outside module, the input variables are the hyperedge
weights $w_e \forall e$ and the outputs are the marginal probabilities
$p_w(i) \forall i$ of each node $i$ in the hypergraph. The latter are
a function of the inside $\beta_i$ and outside $\alpha_j$
probabilities. We initialize $\alpha_{\text{root}} = 1$.
\begin{align}
  \beta_i &= \sum_{e \in I(i)} w_e \prod_{j \in T(e)} \beta_j \\
  \alpha_j &= \sum_{e \in O(i)} w_e \, \alpha_{H(e)} \prod_{j \in T(e) : j \neq i} \beta_j  \\
  p_w(i) &= \alpha_i \beta_i / \beta_{\text{root}} \label{eq:insideoutsidemarginals} 
\end{align}
For each node $i$, we define the set of incoming edges $I(i)$ and
outgoing edges $O(i)$. The antecedents of the edge are $T(e)$, the
parent of the edge is $H(e)$, and its weight is $w_e$.

Below we use the concise notation of an adjoint $\adj{\vc{y}} =
\frac{\partial J}{\partial \vc{y}}$, a derivative with respect to objective $J$.
For the {\bf backward pass} through the inside-outside AD module, the inputs
are $\adj{p_w(i)} \forall i$ and the outputs are $\adj{w_e} \forall
e$.  We also compute the adjoints of the intermediate quantities
$\adj{\beta_j}, \adj{\alpha_i}$.  We first compute $\adj{\alpha_i}$
bottom-up. Next $\adj{\beta_j}$ are computed top-down. The adjoints
$\adj{w_e}$ are then computed in any order.
{\small
\begin{align}
  &\adj{\alpha_i} = \adj{p_w(i)} \deriv{p_w(i)}{\alpha_i}  + \sum_{e \in I(i)} \sum_{j \in T(e)} \adj{\alpha_j} \deriv{\alpha_j}{\alpha_i} \\
  &\adj{\beta_{\text{root}}} = \sum_{i \neq \text{root}} \adj{p_w(i)} \deriv{p_w(i)}{\beta_{\text{root}}} \\
  &\adj{\beta_j} = \adj{p_w(j)} \deriv{p_w(j)}{\beta_j}  + \sum_{e \in O(j)} \adj{\beta_{H(e)}} \deriv{ \beta_{H(e)} }{ \beta_j } \\
  &                       \qquad+ \sum_{e \in O(j)} \sum_{k \in T(e) : k \neq j} \adj{\alpha_k} \deriv{ \alpha_k }{ \beta_j }  \quad \forall j \neq \text{root} %
\end{align}
\begin{align}
  & \adj{w_e} = \adj{\beta_{H(e)}} \deriv{\beta_{H(e)}}{w_e} + \sum_{j \in T(e)} \adj{\alpha_j} \deriv{ \alpha_j }{ w_e }
\end{align}
}
Below, we show the partial derivatives required for the adjoint computations. 
\renewcommand{\deriv}[2]{\frac{\partial #1}{\partial #2}}
{\small
\begin{align*}
&       \deriv{p_w(i)}{\alpha_i} = \beta_i / \beta_{\text{root}}, 
&&&  \deriv{p_w(i)}{\beta_{\text{root}}} = - \alpha_i \beta_i / (\beta_{\text{root}}^2), \\
&       \deriv{p_w(i)}{\beta_i} = \alpha_i / \beta_{\text{root}} &&
\end{align*}
}
\noindent For some edge $e$, let $i = H(e)$ be the parent of the edge and $j, k \in T(e)$ be among its antecendents. 
{\small
\begin{align*}
&         \deriv{\beta_i}{\beta_j} = w_e \prod_{k \in T(e): k \neq j} \beta_k,
\quad  \deriv{\beta_{H(e)}}{w_e} = \prod_{j \in T(e)} \beta_j \\
&         \deriv{\alpha_j}{\alpha_i} = w_e \prod_{k \in T(e) : k \neq j} \beta_k, 
\quad  \deriv{ \alpha_j }{ w_e } = \alpha_{H(e)} \prod_{k \in T(e) : k \neq j} \beta_k \\
&         \deriv{\alpha_k}{\beta_j} = w_e \alpha_H(e) \prod_{l \in T(e) : l \neq j, l \neq k} \beta_l 
\end{align*}
}
This backpropagation method is used for both Figure
\ref{fig:feedforward} C.2 and E.2.

\section{Other Learning Settings}

\paragraph{Loss-aware Training with Exact Inference}
\label{sec:lossawareexactinf}
Backpropagating through inference, decoder, and loss need not
be restricted to \emph{approximate} inference algorithms. \newcite{li-eisner-2009}
optimize Bayes risk with exact inference on a hypergraph for
machine translation. Each of our
differentiable loss functions (\S~\ref{sec:backloss}) can also be
coupled with exact inference. For a first-order parser, BP is
exact. Yet, in place of modules (B), (C), and (D) in Figure
\ref{fig:feedforward}, we can use a standard dynamic programming
algorithm for dependency parsing, which is simply another instance
of inside-outside on a hypergraph (\S~\ref{sec:backinsideoutside}).
The exact marginals from inside-outside
\eqref{eq:insideoutsidemarginals} are then fed forward into the
decoder/loss module (E).

\paragraph{Conditional and Surrogate Log-likelihood}
\label{sec:cll}
The standard approach to training is conditional log-likelihood (CLL)
maximization \cite{smith_eisner_2008_bp}, which does not take inexact
inference into account.
When inference is exact, this baseline computes the true
gradient of CLL. When inference is approximate,
 this baseline uses the approximate marginals from BP in
place of their exact values in the gradient.  {}
The literature refers to
this approximation-\emph{un}aware training method as \emph{surrogate
  likelihood} training since it returns the ``wrong'' model even under
the assumption of infinite training data
\cite{wainwright_estimating_2006}. Despite this, the surrogate
likelihood objective is commonly used to train CRFs. 
CLL and approximation-aware training are not mutually
exclusive. Training a standard factor graph with ERMA and a
log-likelihood objective recovers CLL exactly
\cite{stoyanov_empirical_2011}. 

\section{Experiments}
\label{sec:experiments}

{}

\subsection{Setup}

\paragraph{Features}
\label{sec:features}
As the focus of this work is on a novel approach to training, we look
to prior work for model and feature design.  We add $O(n^3)$
second-order grandparent and arbitrary sibling factors as in
\newcite{riedel_relaxed_2010} and \newcite{martins_turbo_2010}. 
We use standard feature sets for first-order \cite{mcdonald_online_2005} and
second-order \cite{carreras_experiments_2007} parsing. %
Following \newcite{rush_vine_2012}, we also include a version of each
part-of-speech (POS) tag feature, with the coarse POS tags from
\newcite{petrov_universal_2012}. We use feature hashing
\cite{ganchev_small_2008,attenberg2009feature} and restrict to at most 20 million
features. We leave the incorporation of third-order features to future
work.
{}

\paragraph{Pruning}
To reduce the time spent on feature extraction, we enforce the
\emph{type-specific} dependency length bounds from
\newcite{eisner_parsing_2005} as used by \newcite{rush_vine_2012}: the
maximum allowed dependency length for each tuple $($parent tag,
child tag, direction$)$ is given by the maximum observed length for
that tuple in the training data.
Following \newcite{koo_efficient_2010}, we train an (exact)
first-order model {}
and for each token prune any parents for which the
marginal probability is less than 0.0001 times the maximum parent
marginal for that token.\footnote{We expect this to be the least
  impactful of our approximations: \newcite{koo_efficient_2010} report
  $99.92\%$ oracle accuracy for English.} On a per-token basis, we
further restrict to the ten parents with highest marginal probability as
in \newcite{martins_concise_2009}. {} The pruning model uses a simpler 
feature set as in \newcite{rush_vine_2012}.

{}

\paragraph{Data}
We consider 19 languages from the CoNLL-2006
\cite{buchholz_conll-x_2006} and CoNLL-2007 \cite{nivre_conll_2007}
Shared Tasks. We also convert the English Penn Treebank (PTB)
\cite{marcus_building_1993} to dependencies using the head rules from
\newcite{yamada_statistical_2003} (PTB-YM).
We evaluate unlabeled attachment
accuracy (UAS) using gold POS tags for the CoNLL languages, and
predicted tags from
TurboTagger\footnote{\tiny{\url{http://www.cs.cmu.edu/~afm/TurboParser}}} for
the PTB.
Unlike most prior work, we hold out 10\% of each CoNLL training dataset as
development data.

{}

Some of the CoNLL languages contain nonprojective edges. With the
projectivity constraint, {} the model assigns zero probability to such
trees. For approximation-aware training this is not a problem;
{}
however CLL training
cannot handle such trees. For CLL only, we \emph{projectivize} the
training trees following \cite{carreras_experiments_2007} by finding the maximum projective
spanning tree under an oracle model which assigns score +1 to edges in
the gold tree and 0 to the others. {}
We always evaluate on the
nonprojective trees for comparison with prior work.
{}

\begin{figure}[tb]
\centering
\begin{overpic}[width=\linewidth,tics=10]{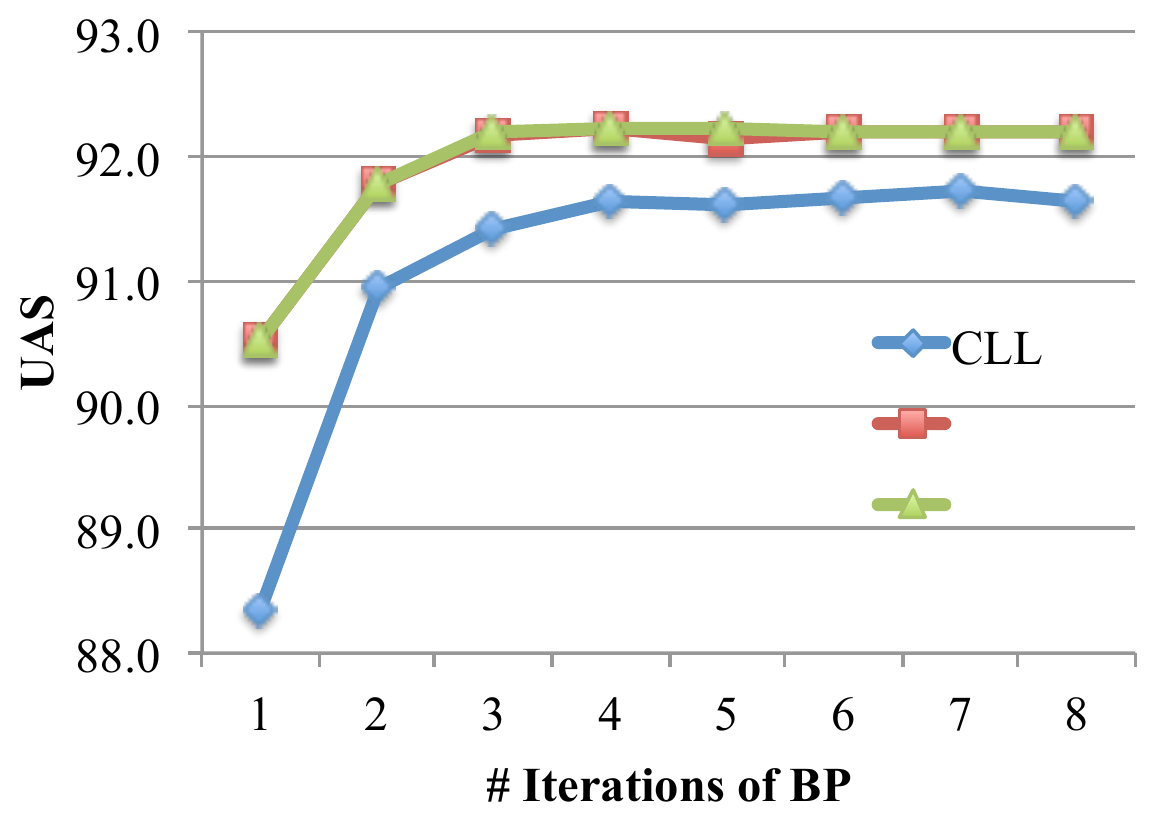}
 \put (82.5,33) {{\small L$_2$}}
 \put (82.5,26.5) {{\small L$_2$+AR}}
\end{overpic}
\caption{Speed accuracy tradeoff of UAS vs. the number of BP
  iterations for standard conditional likelihood training (CLL) and
  our approximation-aware training with either an L$_2$ objective (L$_2$) or
  a staged training of L$_2$ followed by annealed risk (L$_2$+AR). Note that
  x-axis shows the number of iterations used for {\em both} training and
  testing. We use a 2nd-order model with Grand.+Sib. factors.}
\label{fig:uasVsBpIterations}
{}
\end{figure}

\begin{figure}[tb]
\centering
\begin{overpic}[width=\linewidth,tics=10]{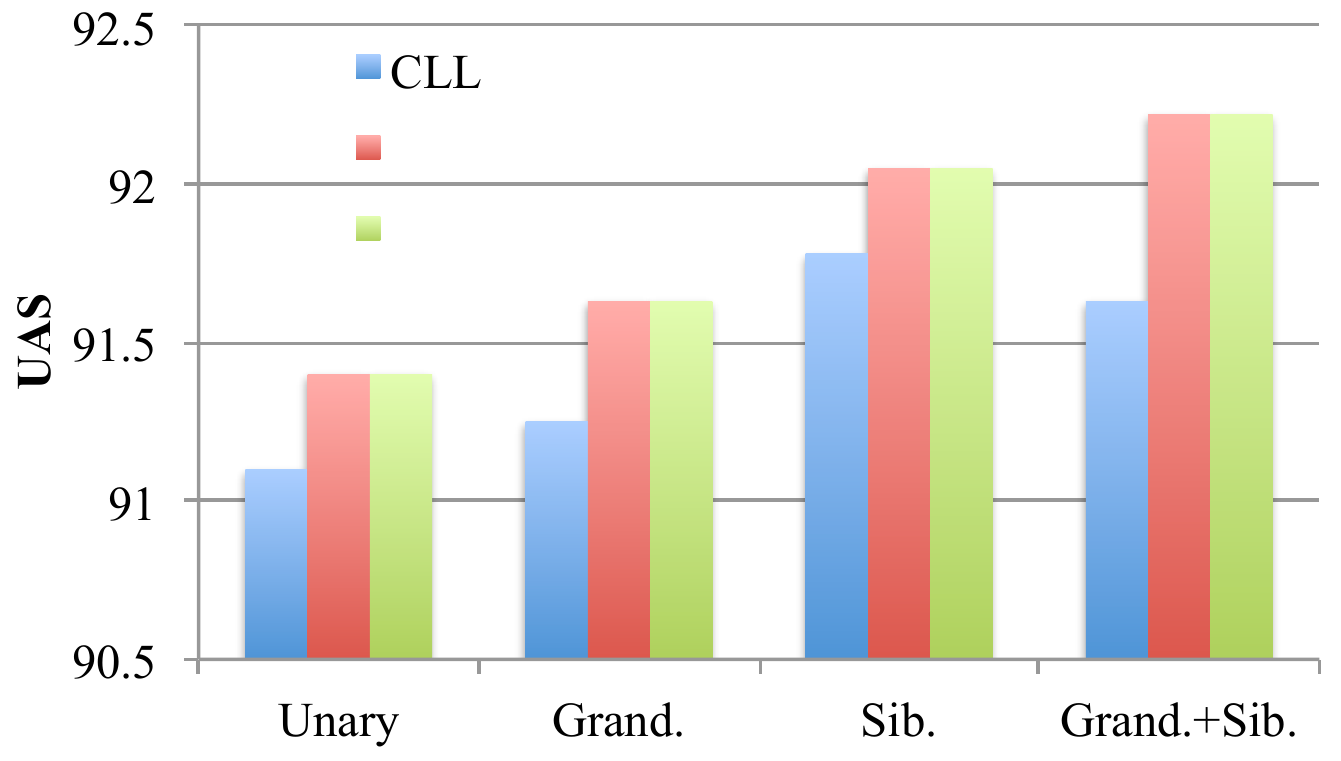}
 \put (30,44.7) {{\footnotesize L$_2$}}
 \put (30,38.5) {{\footnotesize L$_2$+AR}}
\end{overpic}
\caption{UAS vs. the types of 2nd-order factors included in the model
  for approximation-aware training and standard conditional likelihood
  training. All models include 1st-order factors
  (Unary).  The 2nd-order models include grandparents (Grand.),
  arbitrary siblings (Sib.), or both (Grand.+Sib.)---and use 4
  iterations of BP.}
\label{fig:uasVsFactorTypes}
\end{figure}

\paragraph{Learning Settings}
We compare three learning settings. The first, our baseline, is
conditional log-likelihood training (CLL) (\S~\ref{sec:cll}). As is common
in the literature, we conflate two distinct learning settings
(conditional log-likelihood/surrogate log-likelihood) under
the single name ``CLL'' allowing the inference method (exact/inexact) to differentiate
them.
The second learning setting is approximation-aware learning
(\S~\ref{sec:learning}) with either our L$_2$ distance objective (L$_2$) or our layer-wise training
method (L$_2$+AR) which takes the L$_2$-trained model as an initializer
for our annealed risk (\S~\ref{sec:backloss}). 
The annealed risk objective requires an annealing schedule: over the
course of training, we
linearly anneal from initial temperature $T=0.1$ to $T=0.0001$,
updating $T$ at each iteration of stochastic optimization.
The third uses the same two objectives, L$_2$ and L$_2$+AR, but with exact
inference (\S~\ref{sec:lossawareexactinf}).
The $\ell_2$-regularizer weight is $\lambda = \frac{1}{0.1D}$\CommentForSpace{where $D$ is the number of training sentences}. 
Each method is trained by AdaGrad for 10 epochs with early stopping (i.e. the model with the
highest score on dev data is returned). The learning rate for each
training run is dynamically tuned on a sample of the training data.

{}

\subsection{Results}
\label{sec:results}

Our goal is to demonstrate that our approximation-aware training method leads to improved
parser accuracy as compared with the standard training approach of conditional
log-likelihood (CLL) maximization \cite{smith_eisner_2008_bp}, which does not
take inexact inference into account.
The two key findings of our experiments are that our learning approach
is more robust to (1) decreasing the number of iterations of BP and
(2) adding additional cycles to the factor graph in the form of
higher-order factors. In short: our approach leads to faster inference
and creates opportunities for more accurate parsers.
{}

\CommentForSpace{
The goal of our training method is to account for two approximations
made by BP: (1) treating incoming messages as independent even when
the factor graph contains cycles such as those in Figure
\ref{fig:dpFactorGraph} and (2) stopping after a small number of
iterations, before the messages have converged.  Addressing the first
is important if we want to be able to incorporate higher
order factors. The purpose of dependency parsing by belief propagation
is to maintain a low polynomial complexity while incorporating
additional higher order factors (e.g. $O(n^3T)$ for our 2nd-order
factors). However we must still run a first order parser and perform
message computations at each iteration, so addressing the second
approximation matters for speed.
}

\paragraph{Speed-Accuracy Tradeoff}

Our first experiment is on English dependencies. For English PTB-YM,
Figure \ref{fig:uasVsBpIterations} shows accuracy as a function of the
number of BP iterations for our second-order model with both arbitrary sibling
and grandparent factors on English. We find that our training methods
(L$_2$ and L$_2$+AR) obtain higher accuracy than standard training (CLL),
particularly when a small number of BP iterations are used and the
inference is a worse approximation.
Notice that with just \emph{two} iterations of BP, the parsers trained by
our approach obtain accuracy equal to the CLL-trained parser with
\emph{four} iterations.
Contrasting the two objectives for our approximation-aware training,
we find that our simple L$_2$ objective performs very well. In fact, in
only one case at 6 iterations, does the additional annealed risk
(L$_2$+AR) improve performance on test data.  In our development
experiments, we also evaluated AR without using L$_2$ for initialization 
and we found that it performed worse than either of CLL and L$_2$
alone. That AR performs only slightly better than L$_2$ (and not worse)
in the case of L$_2$+AR is likely due to early stopping on dev data,
which guards against selecting a worse model.

\paragraph{Increasingly Cyclic Models}

Figure \ref{fig:uasVsFactorTypes} contrasts accuracy with the type of 
2nd-order factors (grandparent, sibling, or both) included in the
model for English, for a fixed budget of 4 BP iterations. As we add additional higher-order factors, the
model has more loops thereby making the BP approximation more
problematic for standard CLL training. By contrast, our training 
performs well even when the factor graphs have many cycles.

Notice that our advantage is not restricted to the case of loopy
graphs. Even when we use a 1st-order model, for which BP inference is
exact, our approach yields higher accuracy parsers than CLL
training. We postulate that this improvement comes from our
choice of the L$_2$ objective function.  
{}
Note the following subtle point:
when inference is exact, the CLL estimator is actually a special case
of our approximation-aware learner---that is, CLL computes the same
gradient that our training by backpropagation would if we used
log-likelihood as the objective. 
Despite its appealing theoretical justification, the AR
objective that approaches empirical risk minimization in the 
limit consistently provides no improvement over our L$_2$ objective.

\paragraph{Exact Inference with Grandparents}

When our factor graph includes unary and grandparent factors, exact
inference in $O(n^4)$ time is possible using the dynamic programming algorithm for
\emph{Model 0} of \newcite{koo_efficient_2010}. Table
\ref{tab:exactVsApproxGrand} compares four parsers, by considering two
training approaches and two inference methods. The training approaches
are CLL and approximation-aware inference with an L$_2$ objective. The
inference methods are BP with only four iterations or exact inference
by dynamic programming. On test UAS, we
find that both the CLL and L$_2$ parsers with exact inference outperform approximate
inference---though the margin for CLL is much larger. Surprisingly, our L$_2$-trained parser, which uses only 4 iterations 
of BP and $O(n^3)$ runtime, does just as well as CLL with exact
inference. Our L$_2$ parser with exact inference performs the best.

\begin{table}[tb]
  \centering
  {\small
    \begin{tabular}{ll|c|c}
    \tbhead{Train} & \tbhead{Inference} & \tbhead{Dev UAS} & \tbhead{Test UAS} \\
    \hline
    CLL & BP 4 iters & 91.37 & 91.25 \\
    CLL & Exact & 91.99 & 91.62 \\
    L$_2$ & BP 4 iters & 91.83 & 91.63 \\
    L$_2$ & Exact & 91.91 & 91.66 \\
    \hline
    \end{tabular}%
  }
  \caption{The impact of exact vs. approximate inference on a 2nd-order model 
    with grandparent factors only. Results are for the development (\S~22) and test 
    (\S~23) sections of PTB-YM.}
  \label{tab:exactVsApproxGrand}%
\end{table}%

\CommentForSpace{ 
\begin{figure}[tb]
\centering
\includegraphics[width=1.0\linewidth]{fig/uasdiff_vs_bp_iterations.pdf}
\caption{Improvement in unlabeled attachment
  score on test data (UAS) given by using our training method (L$_2$)
  instead of conditional log-likelihood training (CLL) for 19
  languages from CoNLL-2006/2007. The improvements are
calculated directly from the results in Table \ref{tab:conllResults}.}
\label{fig:uasDiffVsBpIters}
\end{figure}
}

\newcommand{\myd}[1]{} %
\newcommand{\myn}[1]{{\color{red}#1}} %
\newcommand{\myp}[1]{{\color{blue}#1}} %
\begin{table*}[t]
  \centering
  {\footnotesize
    \begin{tabular}{c|cc|cc:cc:cc:cc|}
    \multicolumn{1}{c|}{} & \multicolumn{2}{c|}{\tbhead{1st-order}} & \multicolumn{8}{c|}{\tbhead{2nd-order (With  given num. BP iterations)}} \\
    \multicolumn{1}{c|}{} & \multicolumn{2}{c|}{\tbhead{}} & \multicolumn{2}{c:}{\tbhead{1}} & \multicolumn{2}{c:}{\tbhead{2}} & \multicolumn{2}{c:}{\tbhead{4}} & \multicolumn{2}{c|}{\tbhead{8}} \\
    \multicolumn{1}{c|}{\tbhead{Language}} & \tbhead{CLL} & \tbhead{L$_2${\tiny-CLL}} & \tbhead{CLL} & \tbhead{L$_2${\tiny-CLL}} & \tbhead{CLL} & \tbhead{L$_2${\tiny-CLL}} & \tbhead{CLL} & \tbhead{L$_2${\tiny-CLL}} & \tbhead{CLL} & \tbhead{L$_2${\tiny-CLL}} \\
    \hline 
    \tbhead{ar} & 77.63 & \myn{-0.26} & 73.39 & \myp{+2.21} & 77.05 & \myn{-0.17} & 77.20 & \myp{+0.02} & 77.16 & \myn{-0.07} \\
    \tbhead{bg} & 90.38 & \myn{-0.76} & 89.18 & \myn{-0.45} & 90.44 & \myp{+0.04} & 90.73 & \myp{+0.25} & 90.63 & \myn{-0.19} \\
    \tbhead{ca} & 90.47 & \myp{+0.30} & 88.90 & \myp{+0.17} & 90.79 & \myp{+0.38} & 91.21 & \myp{+0.78} & 91.49 & \myp{+0.66} \\
    \tbhead{cs} & 84.69 & \myn{-0.07} & 79.92 & \myp{+3.78} & 82.08 & \myp{+2.27} & 83.02 & \myp{+2.94} & 81.60 & \myp{+4.42} \\
    \tbhead{da} & 87.15 & \myn{-0.12} & 86.31 & \myn{-1.07} & 87.41 & \myp{+0.03} & 87.65 & \myn{-0.11} & 87.68 & \myn{-0.10} \\
    \tbhead{de} & 88.55 & \myp{+0.81} & 88.06 & 0.00  & 89.27 & \myp{+0.46} & 89.85 & \myn{-0.05} & 89.87 & \myn{-0.07} \\
    \tbhead{el} & 82.43 & \myn{-0.54} & 80.02 & \myp{+0.29} & 81.97 & \myp{+0.09} & 82.49 & \myn{-0.16} & 82.66 & \myn{-0.04} \\
    \tbhead{en} & 88.31 & \myp{+0.32} & 85.53 & \myp{+1.44} & 87.67 & \myp{+1.82} & 88.63 & \myp{+1.14} & 88.85 & \myp{+0.96} \\
    \tbhead{es} & 81.49 & \myn{-0.09} & 79.08 & \myn{-0.37} & 80.73 & \myp{+0.14} & 81.75 & \myn{-0.66} & 81.52 & \myp{+0.02} \\
    \tbhead{eu} & 73.69 & \myp{+0.11} & 71.45 & \myp{+0.85} & 74.16 & \myp{+0.24} & 74.92 & \myn{-0.32} & 74.94 & \myn{-0.38} \\
    \tbhead{hu} & 78.79 & \myn{-0.52} & 76.46 & \myp{+1.24} & 79.10 & \myp{+0.03} & 79.07 & \myp{+0.60} & 79.28 & \myp{+0.31} \\
    \tbhead{it} & 84.75 & \myp{+0.32} & 84.14 & \myp{+0.04} & 85.15 & \myp{+0.01} & 85.66 & \myn{-0.51} & 85.81 & \myn{-0.59} \\
    \tbhead{ja} & 93.54 & \myp{+0.19} & 93.01 & \myp{+0.44} & 93.71 & \myn{-0.10} & 93.75 & \myn{-0.26} & 93.47 & \myp{+0.07} \\
    \tbhead{nl} & 76.96 & \myp{+0.53} & 74.23 & \myp{+2.08} & 77.12 & \myp{+0.53} & 78.03 & \myn{-0.27} & 77.83 & \myn{-0.09} \\
    \tbhead{pt} & 86.31 & \myp{+0.38} & 85.68 & \myn{-0.01} & 87.01 & \myp{+0.29} & 87.34 & \myp{+0.08} & 87.30 & \myp{+0.17} \\
    \tbhead{sl} & 79.89 & \myp{+0.30} & 78.42 & \myp{+1.50} & 79.56 & \myp{+1.02} & 80.91 & \myp{+0.03} & 80.80 & \myp{+0.34} \\
    \tbhead{sv} & 87.22 & \myp{+0.60} & 86.14 & \myn{-0.02} & 87.68 & \myp{+0.74} & 88.01 & \myp{+0.41} & 87.87 & \myp{+0.37} \\
    \tbhead{tr} & 78.53 & \myn{-0.30} & 77.43 & \myn{-0.64} & 78.51 & \myn{-1.04} & 78.80 & \myn{-1.06} & 78.91 & \myn{-1.13} \\
    \tbhead{zh} & 84.93 & \myn{-0.39} & 82.62 & \myp{+1.43} & 84.27 & \myp{+0.95} & 84.79 & \myp{+0.68} & 84.77 & \myp{+1.14} \\
    \hline
    \tbhead{Avg.} & 83.98 & \myp{+0.04} & 82.10 & \myp{+0.68} & 83.88 & \myp{+0.41} & 84.41 & \myp{+0.19} & 84.34 & \myp{+0.31} \\
    \hline
    \end{tabular}%
  }
  \caption{Results on 19 languages from CoNLL-2006/2007. For languages appearing in both
  datasets, the 2006 version was used, except for Chinese (\tbhead{zh}). Evaluation follows the 2006
  conventions and excludes punctuation. We report \emph{absolute} UAS for the baseline (CLL) and the \emph{improvement}
 in UAS for L$_2$ over CLL (L$_2${\tiny-CLL}) with \myp{positive}/\myn{negative} differences in blue/red.
 The average UAS and average difference across all languages (\tbhead{Avg.}) is given. \CommentForSpace{These results are further analyzed
  in Figure \ref{fig:uasDiffVsBpIters}.}}
  \label{tab:conllResults}%
\end{table*}%

\paragraph{Other Languages}

{}

Our final experiments evaluate our approximation-aware learning
approach across 19 languages from CoNLL-2006/2007 (Table \ref{tab:conllResults}).
We find that, on average,
approximation-aware training with an L$_2$ objective obtains higher UAS
than CLL training. This result holds for both 1st- and 2nd-order
models with grandparent and sibling factors with 1, 2, 4, or 8
iterations of BP. 
\CommentForSpace{Figure \ref{fig:uasDiffVsBpIters} presents the results of Table
\ref{tab:conllResults} visually.} Table \ref{tab:conllResults} also shows the relative improvement in
UAS of L$_2$ vs CLL training for each language as we vary the maximum
number of iterations of BP. We find that the approximation-aware
training doesn't always outperform CLL training---only in the
aggregate. Again, we see the trend that our training approach
yields more significant gains when BP is restricted to a small number
of maximum iterations.

\section{Discussion}

The purpose of this work was to explore ERMA and related training methods
for models which incorporate structured factors.  We applied these methods
to a basic higher-order dependency parsing model, because that was
the simplest and first \cite{smith_eisner_2008_bp} instance of
structured BP. In future work, we hope to explore further models with
structured factors---particularly those which jointly account for
multiple linguistic strata (e.g. syntax, semantics, and
topic). Another natural extension of this work is to explore other
types of factors: here we considered only exponential-family potential
functions (commonly used in CRFs), but any differentiable function
would be appropriate, such as a neural network. 

Our primary contribution is approximation-aware training for
structured BP. While our experiments only consider
dependency parsing, our approach is applicable for any constraint
factor which amounts to running the inside-outside algorithm on a
hypergraph. Prior work has used this \emph{structured} form
of BP to do dependency parsing \cite{smith_eisner_2008_bp}, CNF grammar parsing \cite{naradowsky_grammarless_2012},
TAG \cite{auli_comparison_2011}, ITG-constraints for
phrase extraction \cite{burkett_fast_2012}, and graphical models over
strings \cite{dreyer_graphical_2009}.  {}  Our training methods could be applied to such tasks as well.

\section{Conclusions}
\label{sec:conclusions}

We introduce a new approximation-aware learning framework for belief
propagation with structured factors. We present differentiable
objectives for both empirical risk minimization (a la. ERMA) and a
novel objective based on L$_2$ distance between the inferred beliefs and
the true edge indicator functions.  Experiments on the English Penn
Treebank and 19 languages from CoNLL-2006/2007 shows that our
estimator is able to train more accurate dependency parsers with fewer
iterations of belief propagation than standard conditional
log-likelihood training, by taking approximations into account. 
Our
code is available in a general-purpose library for structured BP,
hypergraphs, and backprop.\footnote{\url{http://www.cs.jhu.edu/~mrg/software/}}

{}

{}

\bibliographystyle{acl2012}
\bibliography{references}

\end{document}

%% file: feedforward.tex
\tikzstyle{module} = [rectangle,
           rounded corners,
           draw=black, solid, very thick,
           minimum height=2em,
           inner sep=2pt,
           text centered,]
\tikzstyle{zoom} = [rectangle,
           rounded corners=0mm,
           draw=black, dashed,
           minimum height=2em,
           inner sep=2pt,
           text centered,]
\tikzstyle{tinyzoom} = [rectangle,
           rounded corners=0mm,
           draw=black, densely dotted,
           minimum height=2em,
           inner sep=2pt,
           text centered,]
\tikzstyle{empty} = [text centered]
\tikzstyle{line} = [draw, solid, -triangle 45]
\tikzstyle{zoomline} = [draw, dashed]
    
\newcommand{\filler}[3]{
  \begin{tabular}{c}
   \multicolumn{1}{l}{(#1) \textbf{#2}} \\
    #3
  \end{tabular}
}

\newcommand{\dlmodule}[1]{
  \begin{tikzpicture}
    \dlnode{#1}{tinyzoom, scale=0.2}
  \end{tikzpicture}
}
\newcommand{\dlnode}[2]{
\node [#2] (#1) {
    \begin{tikzpicture}
      \node [module] (er) {\filler{E.3}{Expected Recall}{}};
      \node [module, below = of er] (io2) {\filler{E.2}{Inside-Outside}{}};
      \node [module, below = of io2] (anneal) {\filler{E.1}{Anneal Beliefs}{}};
      \path [line] (io2) -- (er);
      \path [line] (anneal) -- (io2);
    \end{tikzpicture}
  };
}

\newcommand{\ptreemodule}[1]{
  \begin{tikzpicture}
    \ptreenode{#1}{tinyzoom, scale=0.2}
  \end{tikzpicture}
}
\newcommand{\ptreenode}[2]{
    \node [#2] (#1) {
      \begin{tikzpicture}
        \node [module] (msgout) {\filler{C.3}{Outgoing Messages}{}}; 
        \node [module, below = of msgout] (io) {\filler{C.2}{Inside-Outside}{}}; 
        \node [module, below = of io] (msgratios) {\filler{C.1}{Message Ratios}{}};
        \path [line] (io) -- (msgout); \path [line] (msgratios) -- (io);
      \end{tikzpicture}
    };
}

\begin{tikzpicture}[remember picture, node distance = 0.5cm, auto, spy using outlines]
  \node [module] (decodeloss) {\filler{E}{Decode and Loss}{
         \raisebox{.5em}{ $\qquad J(\theta; \vc{x}, \vc{y}^*) = $ }\dlmodule{dl1}
    }};
  \node [module, below = of decodeloss] (beliefs) {\filler{D}{Beliefs}{
      $\tilde{b}_i(x_i) = \ldots$,\, $\tilde{b}_{\alpha}(\vc{x_{\alpha}}) =\ldots$
    }};
  \node [module, below = of beliefs] (msgs3)  {\filler{C}{Messages at time $T$}{
      $\tilde{\msg}^{(T)}_{i \rightarrow \alpha}(x_i) =\ldots$,\,
      $\tilde{\msg}^{(T)}_{\alpha \rightarrow i}(x_i) = \ldots$\\
      \raisebox{.5em}{ $\tilde{\msg}^{(T)}_{\PTree{} \rightarrow i}(x_i) = $ } \ptreemodule{ptree3}
    }};
  \node [empty, below = of msgs3] (elipsis3) { $\cdots$ };
  \node [module, below = of elipsis3] (msgs2) {\filler{C}{Messages at time $t$}{
      $\tilde{\msg}^{(t)}_{i \rightarrow \alpha}(x_i) =\ldots$,\,
      $\tilde{\msg}^{(t)}_{\alpha \rightarrow i}(x_i) = \ldots$\\
      \raisebox{.5em}{ $\tilde{\msg}^{(t)}_{\PTree{} \rightarrow i}(x_i) = $ } \ptreemodule{ptree2}
    }};
  \node [empty, below = of msgs2] (elipsis2) { $\cdots$ };
  \node [module, below = of elipsis2] (msgs1) {\filler{C}{Messages at time $t=1$}{
      $\tilde{\msg}^{(1)}_{i \rightarrow \alpha}(x_i) =\ldots$,\,
      $\tilde{\msg}^{(1)}_{\alpha \rightarrow i}(x_i) = \ldots$\\
      \raisebox{.5em}{ $\tilde{\msg}^{(1)}_{\PTree{} \rightarrow i}(x_i) = $ } \ptreemodule{ptree1}
    }};
  \node [module, below = of msgs1, yshift=-0.5cm, xshift=-2cm] (factors) {\filler{A}{Compute Potentials}{
      $\psi_{\alpha}(\vc{x_{\alpha}}) = \exp(\vc{\theta} \cdot \vc{f}(\vc{x_{\alpha}}))$
    }};
  \node [module, right = of factors] (initmsgs) {\filler{B}{Initial Messages}{
      $\msg^{(0)}_{i \rightarrow \alpha}(x_i) = 1$ \\
      $\msg^{(0)}_{\alpha \rightarrow i}(x_i) = 1$
    }};

  \path [line] (beliefs) -- (decodeloss);
  \path [line] (msgs3) -- (beliefs);
  \path [line] (elipsis3) -- (msgs3);
  \path [line] (msgs2) -- (elipsis3);
  \path [line] (elipsis2) -- (msgs2);
  \path [line] (msgs1) -- (elipsis2);
  \path [line] (initmsgs) -- (msgs1);
  \path [line] (factors) edge[bend left=60] (msgs1);
  \path [line] (factors) edge[bend left=50] (msgs2);
  \path [line] (factors) edge[bend left=50] (msgs3);

  \ptreenode{ptree}{zoom, right = of msgs2} 
  \dlnode{dl}{zoom, above = of ptree, yshift = 0.5cm}


  \draw [zoomline] (ptree2.north west) -- (ptree.north west);
  \draw [zoomline] (ptree2.north east) -- ($(ptree.west)+(0,0.7)$);
  \draw [zoomline] (ptree2.south west) -- (ptree.south west);
  \draw [zoomline] (ptree2.south east) -- ($(ptree.west)+(0,-1.0)$); 
  \draw [zoomline] (dl1.north west) -- (dl.north west);
  \draw [zoomline] (dl1.north east) -- (dl.north east);
  \draw [zoomline] (dl1.south west) -- (dl.south west);
  \draw [zoomline] (dl1.south east) -- ($(dl.west)+(0,-1.0)$); 

\end{tikzpicture}